# SRAW-Attack: Space-Reweighted Adversarial Warping Attack for SAR Target Recognition

Yiming Zhang, Weibo Qin, Yuntian Liu, *Student Member, IEEE,* and Feng Wang, *Member, IEEE*

***Abstract*—Synthetic aperture radar (SAR) imagery exhibits intrinsic information sparsity due to its unique electromagnetic scattering mechanisms. Despite the widespread adoption of deep neural network (DNN)-based SAR automatic target recognition (SAR-ATR) systems, they remain vulnerable to adversarial examples and tend to over-rely on background regions, leading to degraded adversarial robustness. Existing adversarial attacks for SAR-ATR often require visually perceptible distortions to achieve effective performance, thereby necessitating an attack method that balances effectiveness and stealthiness. In this paper, a novel attack method termed Space-Reweighted Adversarial Warping (SRAW) is proposed, which generates adversarial examples through optimized spatial deformation with reweighted budgets across foreground and background regions. Extensive experiments demonstrate that SRAW significantly degrades the performance of state-of-the-art DNN-based SAR-ATR models and consistently outperforms existing methods in terms of imperceptibility. The source code is made available at https://github.com/boremycin/SAR-ATR-TransAttack.**

***Index Terms*—Adversarial attack, automatic target recognition, image warping, transferability, synthetic aperture radar (SAR).**

## I. INTRODUCTION

Deep neural network (DNN)-based synthetic aperture radar (SAR) Automatic Target Recognition (SAR-ATR) has become the paradigm of high-resolution SAR image interpretation. However, it is inherently vulnerable to elaborately designed inputs termed adversarial examples, which can induce erroneous predictions [1]. Motivated by this vulnerability, various adversarial attack methods have been proposed for optical images [2]-[5]. In the SAR domain, related studies [6], [7] have begun to incorporate SAR-specific characteristics, such as microwave-based imaging mechanisms and clutter-dominated grayscale representations, into attack strategies. Nevertheless, how to design attack mechanisms that are both effective and well-aligned with the unique properties of SAR imagery remains an open problem.

SAR imagery has been confirmed to exhibit pronounced background correlation [8], [9], whereby clutter regions with limited semantic content exert a non-negligible influence on SAR-ATR predictions. Belloni et al. [9] further analyzed the interplay among different components, while Li et al. [8] pointed out that biases in SAR data introduce non-causal correlations, encouraging DNNs to overfit background clutter. We refer to the resulting phenomenon as negative robustness, i.e., the models achieve apparent robustness by over-relying on background contextual cues rather than target-specific scattering features. When the DNNs over-parameterized for optical imagery are applied to SAR data, the relatively monotonic background can induce a form of negative robustness. As illustrated in Fig. 1, Grad-CAM [10] reveals that the model's attention is more concentrated on background regions than on the target foreground.

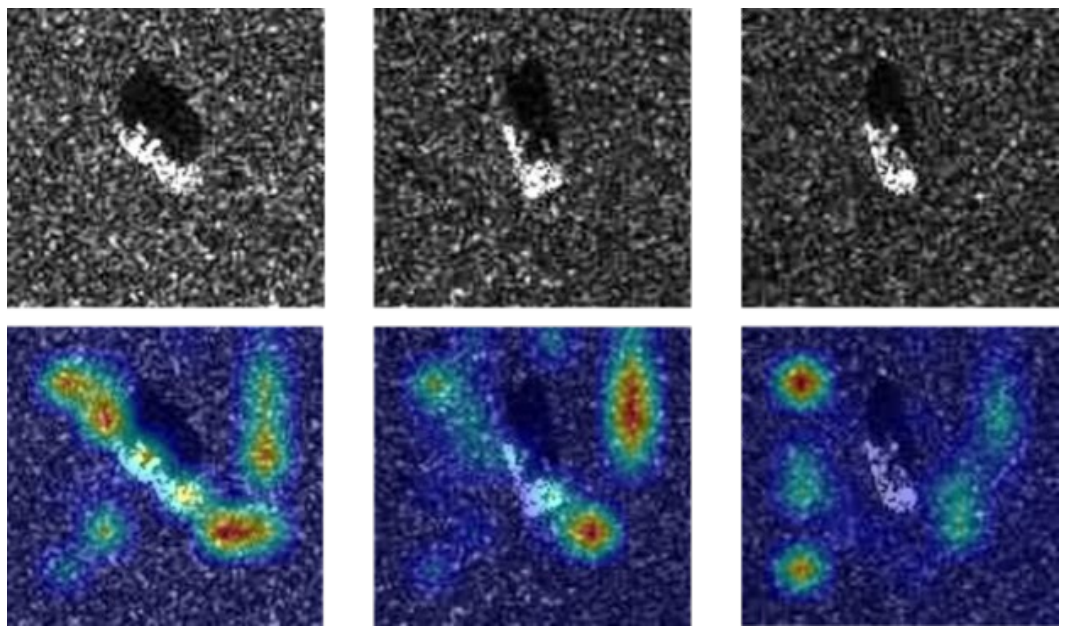

**Fig. 1.** Grad-CAM–based [10] attention heatmap visualization of SAR images, highlighting the discriminative regions exploited by the DNN for target recognition.

In contrast to optical imagery, where adversarial examples are typically imperceptible, the intrinsic information sparsity of SAR data, caused by the electromagnetic scattering mechanism [8], [11], often necessitates more perceptible perturbations to achieve effective attacks, thereby severely undermining their stealthiness. Since the background perturbations are generally less perceptible, a natural strategy is to impose heterogeneous perturbation budgets between foreground and background, assigning aggressive alterations in the background while preserving target consistency. In addition, transformation-based attack methods [3], [12] have been proven to boost both attack effectiveness and transferability. Meanwhile, SAR imagery is sensitive to geometric structure, and subtle topological deformations can shift key scatter centers [7]. Based on the above analysis, introducing spline-based warping [13] mechanisms into adversarial generation exhibits significant potential for SAR-ATR, as it enables effective manipulation of background-dependent cues while maintaining a balance between attack effectiveness and perceptual imperceptibility.

In this letter, a novel adversarial method for SAR imagery is proposed, termed **Space-Reweighted Adversarial Warping (SRAW)**, which generates adversarial examples via Thin-

The computations in this research were performed using the CFFF platform of Fudan University. *(Corresponding author: Feng Wang).*

Yiming Zhang, Weibo Qin, Yuntian Liu, and Feng Wang are with the Discipline and Technology Center of Microwave Vision Intelligent Sensing, School of Information Science and Technology, Fudan University (FDU), Shanghai, 200433, China (e-mail: zhangym21@m.fudan.edu.cn, wbqin21@m.fudan.edu.cn, fengwang@fudan.edu.cn)

Plate Splines (TPS) [13] spatial warping. Rather than employing TPS as an auxiliary image transformation for perturbation optimization, as in existing TPS-based attacks such as DeCoWa [3], SRAW reformulates adversarial example generation as an optimization over warping control points. These control points are updated through gradient-based iterations under space-reweighted deformation constraints between foreground and background regions, enabling an effective and stealthy adversarial configuration.

The framework of our SRAW method is shown in Fig. 2, and the main contributions of this letter are as follows:

1) We identify a fundamental dilemma of adversarial attacks for SAR imagery, where the intrinsic information sparsity of SAR data causes effective attacks to rely on stronger and more perceptible perturbations, making it difficult to simultaneously achieve effectiveness and stealthiness.
2) A novel SAR-specific attack termed Space-Reweighted Adversarial Warping (SRAW) is proposed. Unlike existing TPS-based attacks that employ TPS as an auxiliary image transformation, SRAW directly optimizes TPS control points through backpropagation with a tailored foreground-background reweighting strategy, reformulating adversarial example generation as a spatial deformation optimization problem.
3) Extensive experiments on the MSTAR dataset demonstrate that SRAW consistently outperforms existing established and state-of-the-art (SOTA) attacks in terms of attack success rate while preserving remarkably improved imperceptibility.

## II. Methodology

### A. Preliminaries and Problem Formulation

The input benign image is denoted as $x \in \mathbb{R}^{w\times h}$ with its class label $y$. A DNN model $f(\cdot)$ serves as a SAR-ATR classifier, where $f(x;\theta) = y$, and $\theta$ represents the learned parameters of the target model. As for perturbation-adding attacks, the goal is to craft an adversarial example $x^{adv} = x + \delta$ with the elaborately designed perturbation $\delta$ to mislead the target model into making incorrect predictions, i.e., $f(x^{adv};\theta) \neq y$. To guarantee that the adversarial example remains imperceptible, $\delta$ is limited to a bounded $\ell_p$-norm with the perturbation budget $\epsilon$. Following the widely used setting, the perturbation is constrained by the $\ell_\infty$-norm in this paper. Accordingly, the generation of adversarial examples can be formulated as the following optimization problem:

$$\arg\max_{x^{adv}} \mathcal{L}(f(x^{adv}), y), \qquad s.t.\ \|\delta\|_\infty \leq \epsilon, \tag{1}$$

where $\mathcal{L}(\cdot)$ is the cross-entropy loss, which is commonly applied to DNN models.

Nonetheless, it is arduous to directly manipulate the target model as in (1), and the more practical implementation is to iteratively produce adversarial examples via a surrogate model $f_{sur}(\cdot)$ and mislead the target model through the transferability of adversarial examples, referred to as the black-box attack setting. Hence, the $(t+1)$-th iteration can be formulated as:

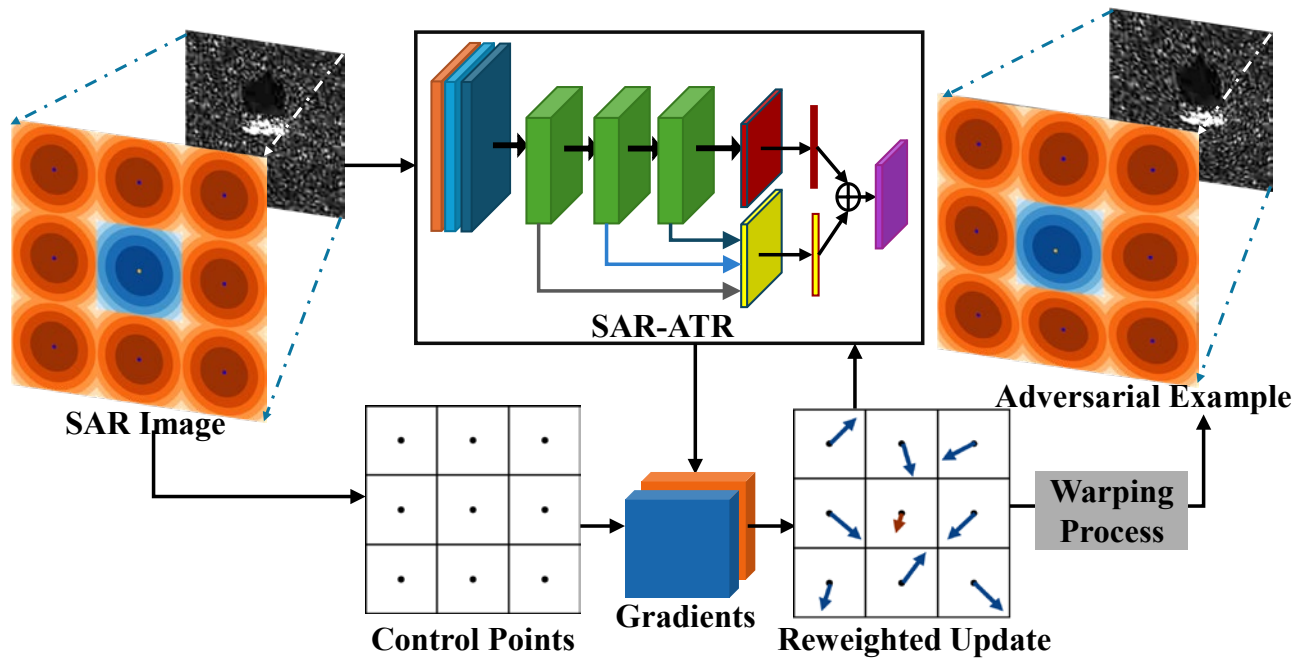

**Fig. 2.** Illustration of the proposed SRAW method. The input SAR image is partitioned into foreground-background regions, where tighter constraints are imposed on the foreground and looser bounds are assigned to the background. Gradients are iteratively computed to optimize the control-point offsets during the reweighted updates, followed by the warping process to generate the adversarial example.

$$x_{t+1}^{adv} = Clip_\epsilon\{x_t^{adv} + \alpha \cdot sign(\nabla_{x_t^{adv}} \mathcal{L}(f_{sur}(x_t^{adv}, y)))\}, \tag{2}$$

where $Clip_\epsilon(\cdot)$ denotes the $\ell_\infty$-norm constraint that restricts the perturbation within an $\epsilon$-radius; $\alpha$ is the iterative step size; $sign(\cdot)$ denotes the sign function and $\nabla_{x_t^{adv}}\mathcal{L}(\cdot)$ represents the gradient of the loss with respect to the adversarial example at iteration $t$.

### B. TPS-Based Spatial Deformation

Considering the trade-off between attack effectiveness and concealment, an initial motivation is to convert the generating paradigm by replacing additive perturbation with spatial deformation. According to the core of the TPS warping algorithm [13], the spatial deformation consists of two main steps: solving the warping parameters and performing the image interpolation. The first step requires two sets of control-point coordinates, i.e.,

$$\boldsymbol{\Gamma}_{src} = \{(u_i, v_i)\}_{i=1}^{N}, \qquad \boldsymbol{\Gamma}_{tar} = \{(u_i', v_i')\}_{i=1}^{N}, \tag{3}$$

where $\boldsymbol{\Gamma}_{src}$ and $\boldsymbol{\Gamma}_{tar}$ denote the source and target control-point sets respectively, and $(u_i, v_i)$ or $(u_i', v_i')$ is the $i$-th point among the set of $N$ control points.

To ensure a smooth interpolation of the deformation effect, the TPS algorithm utilizes the radial basis function:

$$U(r) = r^2 ln(r). \tag{4}$$

This function quantifies the geometric deformation between $\boldsymbol{\Gamma}_{src}$ and $\boldsymbol{\Gamma}_{tar}$. Given that planar coordinate mapping can be decomposed into two separate scalar fields, the spline interpolation functions can be written as:

$$g_{u'}(u,v) = a_1 + a_u u + a_v v + \sum_{i=1}^{N} w_i U(\|\boldsymbol{P}_i - (u,v)\|), \tag{5}$$

$$g_{v'}(u,v) = b_1 + b_u u + b_v v + \sum_{i=1}^{N} w_i' U(\|\boldsymbol{P}_i - (u,v)\|), \tag{6}$$

where $a_1$, $a_u$, $a_v$ and $b_1$, $b_u$, $b_v$ are the affine coefficients, $w_i$ and $w_i'$ are the non-affine kernel weights associated with control point $\boldsymbol{P}_i = (u_i, v_i)$. The coefficients can be solved through the following linear system:

$$\begin{bmatrix} \boldsymbol{K} & \boldsymbol{\Gamma}_{src} \\ \boldsymbol{\Gamma}_{src}^T & \boldsymbol{O} \end{bmatrix} \times \begin{bmatrix} \boldsymbol{W} \\ \boldsymbol{a} \end{bmatrix} = \begin{bmatrix} \boldsymbol{\Gamma}_{tar} \\ \boldsymbol{O} \end{bmatrix}, \tag{7}$$

where $\boldsymbol{K}$ is the kernel matrix defined by $K_{ij} = U(\|\boldsymbol{P}_i - \boldsymbol{P}_j\|)$, $\boldsymbol{w}$ and $\boldsymbol{a}$ denote the non-affine and affine TPS coefficients, respectively.

Given the solved TPS parameters, image interpolation is performed by remapping input pixels according to (5) and (6), using bicubic interpolation. The overall deformation applied to the input $x$ to yield warped output $x'$ can be formulated as the transformation function $\mathcal{T}(\cdot)$ defined as follows:

$$x' = \mathcal{T}(x, \boldsymbol{\Gamma}_{src}, \boldsymbol{\Gamma}_{tar}). \tag{8}$$

From the perspective of adversarial equivalence to perturbation-adding attacks, the warped image $x'$ is directly optimized through the target loss. By applying the chain rule of differentiation, the gradient can be written as:

$$\nabla_{\Gamma}\mathcal{L} = \frac{\partial \mathcal{L}}{\partial x'} \cdot \frac{\partial \mathcal{T}(x, \boldsymbol{\Gamma})}{\partial \boldsymbol{\Gamma}}. \tag{9}$$

The spline-based formulation in (5)-(6) parameterizes spatial deformation on a low-dimensional smooth manifold, constraining the search space and stabilizing the optimization.

### *C. Proposed SRAW Attack*

To capitalize on the negative robustness of DNN-based SAR-ATR and enhance the imperceptibility of the resulting adversarial examples, we propose a **Space-Reweighted Adversarial Warping (SRAW)** attack. The method integrates TPS-based spatial deformation into a momentum-based gradient-driven framework, thereby reformulating adversarial example generation as a spatial deformation optimization problem.

More precisely, the key factor governing the deformation behavior is the offsets $\boldsymbol{\xi}$ between $\boldsymbol{\Gamma}_{src}$ and $\boldsymbol{\Gamma}_{tar}$ when $\boldsymbol{\Gamma}_{src}$ is initially specified, which can thus be expressed as:

$$\boldsymbol{\xi} = \boldsymbol{\Gamma}_{tar} - \boldsymbol{\Gamma}_{src}. \tag{10}$$

In the SRAW attack, the source control-point set $\boldsymbol{\Gamma}_{src}$ is fixed by a priori spatial mesh partition, where the center of each grid cell is selected as a source control point. In this work, a $3 \times 3$ mesh is adopted, yielding nine control points and naturally partitioning the image into one central foreground region and eight surrounding background regions. A coarser mesh cannot explicitly separate the target from the surrounding background, whereas a finer mesh introduces unnecessary deformation degrees of freedom without providing additional benefits. The optimization process can be restricted to the control-point offsets $\boldsymbol{\xi}$. According to the formulations in (1) and (8), the adversarial example derived from the input $x$ is expressed as a spatially warped instance:

$$x^{adv} = \mathcal{T}(x, \boldsymbol{\xi}), \tag{11}$$

where $\mathcal{T}(\cdot)$ is parameterized by $\boldsymbol{\xi}$. Consequently, the adversarial example generation can be cast as the following optimization problem:

$$\arg\max_{\boldsymbol{\xi}} \mathcal{L}\left(f(\mathcal{T}(x, \boldsymbol{\xi})), y\right). \tag{12}$$

Due to the complex TPS mapping relationship and the dimensionality reduction in control-point space, gradients obtained from a single backpropagation step may be noisy and deviate from the true descent direction. To obtain a more reliable update, a multi-warping averaging strategy is adopted. At the $(i+1)$-th iteration, the aggregated gradient $\bar{g}_{i+1}$ is computed as:

$$\bar{g}_{i+1} = \frac{1}{N_w}\sum_{j=1}^{N_w} g_j = \frac{1}{N_w}\sum_{j=1}^{N_w} \nabla_{\boldsymbol{\xi}}\mathcal{L}\left(f(\mathcal{T}(x_i^{adv}, \boldsymbol{\xi}_i)), y\right), \tag{13}$$

where $N_w$ denotes the number of warped samples used for gradient averaging, and $g_j$ is the gradient associated with the $j$-th warped sample.

A momentum-based gradient update is adopted to stabilize optimization. At the $(i+1)$-th iteration, the gradient $g'_{i+1}$ is constructed by combining the current gradient $g_{i+1}$ with the accumulated momentum $g'_i$. The momentum term suppresses gradient noise and promotes a consistent descent direction, resulting in smoother updates and faster convergence. The update rule is given by:

$$g'_{i+1} = \mu \cdot g'_i + \frac{\bar{g}_{i+1}}{\|\bar{g}_{i+1}\|}, \tag{14}$$

where $\mu$ represents the decay factor.

To preserve the visual plausibility of $x^{adv}$, the magnitude of spatial deformation is constrained. The control-point offsets $\boldsymbol{\xi}$ are decomposed as $\boldsymbol{\xi} = \{\boldsymbol{\xi}_{fg}, \boldsymbol{\xi}_{bg}\}$. A tight displacement budget is imposed on $\boldsymbol{\xi}_{fg}$ to preserve target integrity, whereas a looser bound is applied to $\boldsymbol{\xi}_{bg}$, enabling stronger deformation to disrupt the reliance on contextual cues. The update of the control-point offsets at iteration $i+1$ follows a projected gradient step:

$$\boldsymbol{\xi}_{i+1} = \Pi_C(\boldsymbol{\xi}_i + \alpha \cdot g'_i), \tag{15}$$

where $\Pi_C(\cdot)$ is the projection onto the spatial-reweighted feasible set $C$ determined by the region-dependent displacement limits $r_{fg}$ and $r_{bg}$. Specifically,

$$C = \{\boldsymbol{\xi}: \|\boldsymbol{\xi}_{fg}\| \le r_{fg}, \|\boldsymbol{\xi}_{bg}\| \le r_{bg}\}, \tag{16}$$

and α is the step size.

Table I

Classification Accuracy (%) of Target Models

| Model | Training Accuracy | Test Accuracy |
|---|---|---|
| DenseNet [21] | 100.00 | 99.65 |
| VGG [22] | 100.00 | 98.16 |
| ResNet [23] | 100.00 | 98.26 |
| ResNext [24] | 99.16 | 96.22 |
| PyramidNet [25] | 99.93 | 97.16 |
| ViT [26] | 100.00 | 98.06 |
| ConvNeXt [27] | 99.65 | 99.40 |
| Swin-T [28] | 99.95 | 98.51 |

## III. Experiments

### *A. Experimental Setup and Evaluation Measurement*

In this paper, the MSTAR [14] dataset published by the Defense Advanced Research Projects Agency (DARPA) is adopted. The dataset comprises SAR images across ten distinct vehicle categories and is randomly partitioned into training and test sets with a 3:1 ratio in our experiments, with the test set additionally employed for adversarial attack evaluation.

Table II

WHITE-BOX ASR (%) RESULTS OF TARGET MODELS

| Method | DenseNet | VGG | ResNet | ResNext | PyramidNet | ViT | ConvNeXt | Swin-T | Average |
|---|---|---|---|---|---|---|---|---|---|
| FGSM [4] | 60.89 | 79.11 | 64.51 | 83.48 | 58.83 | 71.76 | 65.43 | 73.97 | 69.75 |
| MI-FGSM [2] | 63.97 | 73.36 | 74.94 | 82.42 | 62.00 | 72.28 | 62.86 | 71.36 | 70.40 |
| PGD [15] | 66.03 | 78.06 | 79.13 | 81.87 | 66.22 | 72.28 | 60.27 | 79.70 | 72.95 |
| CW [5] | 62.43 | 75.45 | 68.69 | 85.08 | 60.42 | 66.53 | 63.37 | 75.01 | 69.62 |
| LoRa-PGD [18] | 65.01 | 80.14 | 69.20 | 74.41 | 68.86 | 66.01 | 64.41 | 71.36 | 69.93 |
| DeCoWa [3] | 67.58 | 76.49 | 76.51 | 80.81 | 65.16 | 69.67 | 62.86 | 69.28 | 71.05 |
| BSR [12] | 70.67 | 78.58 | 73.11 | 76.55 | 69.92 | 75.42 | 68.53 | 70.33 | 72.89 |
| MEFA [19] | 67.58 | 77.54 | 78.08 | 81.87 | 70.97 | 70.19 | 69.04 | 77.62 | 74.11 |
| SRAW | **71.18** | **81.19** | **81.73** | **85.61** | **72.03** | **80.65** | **70.59** | **85.94** | **78.62** |

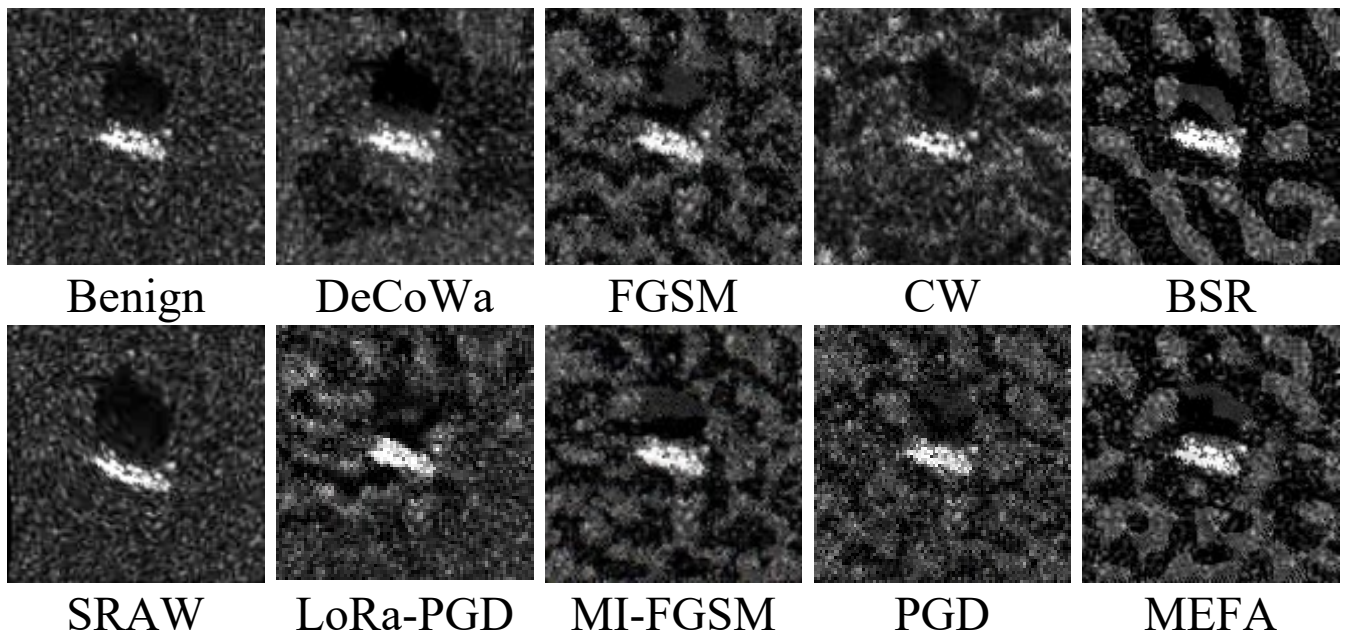

**Fig. 3.** Visual comparison between the benign image and corresponding adversarial examples of other attack approaches.

Table III

ASR (%) RESULTS UNDER ADVERSARIAL DEFENSES

| Method | AT | RS | JC |
|---|---|---|---|
| FGSM | 45.73 | 63.99 | 61.38 |
| MI-FGSM | 49.90 | 71.29 | 43.11 |
| PGD | 48.33 | 77.04 | 46.24 |
| CW | 48.85 | 65.03 | 42.59 |
| LoRa-PGD | 45.73 | 70.77 | 48.33 |
| DeCoWa | 43.64 | 77.04 | 73.91 |
| BSR | 48.85 | 75.99 | 69.20 |
| MEFA | 50.42 | 67.65 | 39.99 |
| SRAW | **62.95** | **79.65** | **78.60** |

To demonstrate the effectiveness and general applicability of SRAW, it is evaluated on a set of representative DNNs with different backbones, including CNN and Transformer. The classification performance of these models is summarized in Table I, indicating strong accuracy on the SAR-ATR task.

For a fair and comprehensive comparison, our method is evaluated against widely used benchmark and recent SOTA adversarial attacks. The attack performance is compared in terms of Attack Success Rate (ASR), which is defined using the indicator function $I(\cdot)$ as:

$$ASR = \frac{1}{N_c}\sum_{i=1}^{N_t} I\left(f(x_i) = y_i \wedge f\left(x_i^{adv}\right) \neq y_i\right), \quad (17)$$

where $N_t$ represents the number of samples in the test set, and $N_c$ is the number of correctly classified benign samples.

*B. Attack Performance Evaluation*

White-box attack experiments are performed independently for each model and each method to compare their attack effectiveness. In this scenario, the target model is used to simultaneously generate and test the same adversarial examples. As reported in Table II, the proposed SRAW attack outperforms all other approaches, achieving an average ASR of **78.62%** across all target models.

Table IV

PERCEPTUAL QUALITY EVALUATION

| Method | PSNR (dB) ↑ | SSIM↑ | LPIPS↓ |
|---|---|---|---|
| FGSM | 31.24 | 0.4567 | 0.3804 |
| MI-FGSM | 30.85 | 0.4369 | 0.4051 |
| PGD | **31.48** | 0.5109 | 0.3440 |
| CW | 29.01 | 0.5633 | 0.3521 |
| LoRa-PGD | 28.91 | 0.5251 | 0.4716 |
| DeCoWa | 26.86 | 0.4715 | 0.3829 |
| BSR | 30.66 | 0.4611 | 0.4255 |
| MEFA | 29.76 | 0.4885 | 0.3979 |
| SRAW | <u>30.09</u> | **0.5724** | **0.2930** |

To further evaluate the adversarial performance of SRAW under more rigorous defensive conditions, ResNet is selected as the target model, and three representative defense strategies, namely PGD-based adversarial training (AT) [15], randomized smoothing (RS) [16] and JPEG compression (JC) [17] are employed. As reported in Table III, while these defenses lower all attacks' efficacy, SRAW consistently achieves the highest ASR under all defense settings, reaching **62.95%**, **79.65%**, and **78.60%** against AT, RS, and JC, respectively. These results manifest the superior robustness of SRAW in realistic defensive scenarios.

*C. Analysis and Discussion*

The visualization of the benign image and corresponding adversarial examples in Fig. 3 demonstrate that SRAW achieves higher visual imperceptibility than other competing approaches. To precisely quantify the perceptual impact of image deformation, three widely used metrics are adopted: Peak Signal-to-Noise Ratio (PSNR), Structural Similarity Index (SSIM), and Learned Perceptual Image Patch Similarity (LPIPS) [20]. As reported in Table IV, SRAW achieves the **highest SSIM** and the **lowest LPIPS** among all compared methods, while maintaining competitive PSNR, indicating superior structural preservation and perceptual invisibility. The overall results demonstrate that SRAW provides a more favorable perceptual trade-off, yielding visually subtle yet effective adversarial examples.

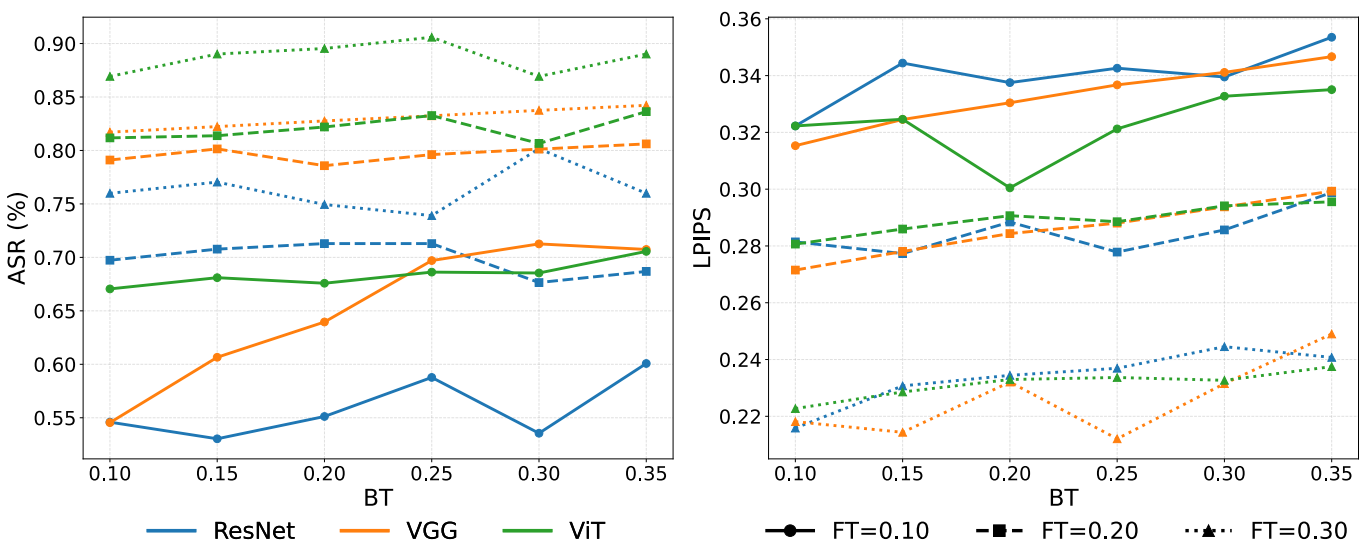

(a) (b)

**Fig. 4.** Parameter sensitivity analysis of SRAW under different foreground thresholds (FT) and background thresholds (BT). (a) ASR. (b) LPIPS. Results are reported on ResNet, VGG, and ViT.

To investigate the impact of the space-reweighted deformation budgets, we vary the foreground threshold (FT) and background threshold (BT) and evaluate the resulting ASR and LPIPS on ResNet, VGG and ViT. The results in Fig. 4 reveal that increasing FT and BT generally improves ASR while increasing LPIPS, indicating a clear trade-off between attack effectiveness and visual imperceptibility. This further demonstrates that the proposed method is not highly sensitive to small perturbations in FT and BT. Although the optimal parameter combination varies slightly, similar sensitivity trends are observed across different models. Accordingly, moderate FT and BT achieve a favorable balance between effectiveness and imperceptibility.

Due to the complicated imaging geometry and unique electromagnetic scattering mechanisms of SAR systems, physically realizing the proposed spatial deformation remains challenging. Deceptive jamming may provide a potential pathway for future real-world feasibility. Moreover, SRAW is most effective when foreground and background regions can be clearly distinguished, while more adaptive space partition strategies deserve further investigation for complex scenes.

## IV. Conclusion

In this letter, we contend that the intrinsic information sparsity of SAR imagery induces a form of negative robustness in DNN-based SAR-ATR systems, leading to excessive reliance on background regions. Motivated by this insight, a novel attack method termed Space-Reweighted Adversarial Warping (SRAW) is proposed, which reformulates adversarial generation as the optimization of spatial warping with reweighted budgets on foreground and background regions.

Comprehensive experiments corroborate that SRAW consistently outperforms existing methods in terms of attack effectiveness and imperceptibility, while remaining robust under adversarial defense mechanisms. Future work will explore the physical feasibility of SRAW by integrating it into the SAR signal formation and processing pipelines.